\documentclass[11pt]{article}
\usepackage[T1]{fontenc}
\usepackage[utf8]{inputenc}
\usepackage{lmodern}
\usepackage[a4paper,margin=1in]{geometry}
\usepackage{amsmath,amssymb}
\usepackage{booktabs}
\usepackage{graphicx}
\usepackage{xurl}
\usepackage[colorlinks=true,allcolors=blue]{hyperref}
\title{Time and Space: A Dual-Axis, Evidence-Grounded Context Substrate for Personal Agents}
\author{\includegraphics[height=0.46cm]{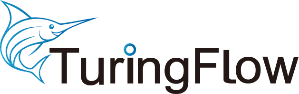}\\[0.4em]
\small Haitao Cheng, Jerry Hu, Longdong Mao, Dong Huang, Dongming Li, Lijuan Zhang, Mengting Zhang}
\date{}
\begin{document}
\maketitle
\begin{abstract}

We argue that a binding constraint for personal agents is not model capacity but the form in which context is represented: the agent does not know which entity a phrase denotes, when a fact held, or which source record licenses a claim. We call the missing representation \textit{spatio-temporal context} (STC) and define it on two axes. The temporal axis carries event time, validity intervals and ingestion time. The spatial axis is a relational space in which entities are points, typed relations are edges, and distance is semantic rather than geographic; physical places are one entity type among many. Every relation is anchored to immutable source records, and every answer must be closed over those records. We formalise the substrate and six invariants, a context operator that composes temporal intersection with relational expansion, an ambiguity-preserving resolution state machine, a conflict detector that converts mutually exclusive facts into user adjudication rather than silent overwrite, and a four-regime consumption taxonomy spanning query, insight, triage and proactive reminder. We describe a local-first prototype of the retrieval layer built on SQLite with on-device embeddings and hybrid lexical-dense recall, report measurements from a deployed store of 341 documents, 340,079 chunks and 190,271 embedded vectors, and specify, but do not yet implement, the analytic layer on which the framework depends. We close with an evaluation protocol with per-invariant metrics for evidence coverage, temporal correctness, conflict precision and interruption cost. The proposal targets the representation of context rather than the volume of it.

\end{abstract}

\vspace{4pt}\noindent\textbf{Keywords:} spatio-temporal context; agent memory; temporal knowledge graphs; provenance; bitemporal modelling; personal agents; evidence-grounded generation; local-first computing\par

\section{Introduction}

A user asks an agent: "When my mother was last in hospital, where was I?" A model can parse the question, but the evidence needed to answer it is spread across artefacts that no single retrieved chunk covers. The relevant facts are scattered across a family chat that calls someone "mom", a bank transfer whose memo also says "mom", a photograph whose reverse side names a mother-in-law, and a hospital registration slip that lists a patient and an accompanying person. No single fragment is sufficient, and each fragment is ambiguous in isolation. The obstacle is not reasoning power but the absence of a representation that ties fragments to the same entities, places them on a timeline, and records which fragment supports which claim.

This is the situation most personal agents are in. Their inputs are heterogeneous artefacts produced by other applications: chat logs, order confirmations, bank statements, photographs with EXIF metadata, travel itineraries, calendar entries, local documents. Retrieval-augmented generation treats these artefacts as text chunks and asks which chunks look similar to the query. That works for document question answering, where the answer usually sits inside one passage. This is often insufficient for questions that require combining facts across sources, resolving identity across surface forms, and respecting the time at which each fact held.

We propose to make this structure explicit. We call the resulting representation \textit{spatio-temporal context} and define it on two axes. The temporal axis records when something happened, the interval over which it remained true, and when the system learned it. The spatial axis is a relational space: entities are points, typed relations are edges, and proximity is measured by relational and semantic similarity rather than by metres. A hospital and a hometown are points of the same kind as a person; geography becomes one attribute of an entity rather than the definition of space. The two axes compose into a single retrieval operator: a context for a query is the set of evidence-backed relations that are simultaneously reachable from the query's anchors in the relational space and valid within the query's temporal focus.

The motivation for the spatial framing comes from the models we are trying to help. Inside a language model, tokens are projected into a high-dimensional vector space, and meaning emerges from relations among positions in that space rather than from the positions alone. Flat-text memory discards precisely this structure: it keeps the tokens and drops the relations. We take this as motivation rather than as support: an explicit relational space is meant to give the agent structure it can cite, not structure it already has.

We make four contributions. (1) A specification of a spatio-temporal context for personal data, with six invariants (Sections 3 and 4). (2) Operators over that representation: a context operator, an ambiguity-preserving consistency test, a functional-exclusivity conflict detector, an evidence-closure rule and a trigger rule (Section 5). (3) A prototype of the retrieval layer, measured on a deployed store (Section 7). (4) An evaluation protocol with per-invariant metrics, none of whose components has been run (Section 9). The analytic layer that the operators presuppose is specified here rather than implemented, and we say so wherever it matters. We propose these as a unit of design and give the metrics by which the proposal can be tested; the framework is the object of the paper, not a result.

\section{Related Work}

\subsection{Memory for language agents}

Memory research for language agents has converged on two families: context-window management and external stores. Packer et al. (2023) treat the context window as a managed resource and page information between tiers, which addresses capacity but not truth. Park et al. (2023) record observations in a memory stream and retrieve by recency, importance and relevance; the stream is timestamped, but a timestamp is an attribute of a memory rather than a validity interval for a fact, and the design does not maintain an entity graph.

Personal knowledge graphs are an older framing of the same problem (Balog \& Kenter, 2019); the entity-centric view they propose is the one we inherit. The temporal knowledge graph family goes further. Rasmussen et al. (2025) build agent memory on a temporally aware graph with bi-temporal edges, distinguishing when a fact was true from when it was recorded, and attach episodes as provenance; their evaluation emphasises conversational memory benchmarks. Niksarli and Baheti (2026) give a graph-native bitemporal store with closed-open intervals for valid and transaction time. Wan et al. (2026) make provenance mandatory: typed fact events carry source spans, and mention time, occurrence time and validity are stored separately, with supersession and contradiction edges and hybrid lexical-dense indexing. Shu et al. (2026) represent episodic memory as time-scoped gists and facts. Lu et al. (2026) maintain a confidence-aware temporal property graph with non-destructive updates. Lupa\c{s}cu and Lupascu (2026) attach event, ingestion and update timestamps to every assertion and type the edges that relate competing assertions. Su et al. (2026) move beyond dialogue time by constructing a semantic timeline over events, so that facts can be retrieved by when they occurred rather than only by when they were said. Yang et al. (2026) survey this space and discuss bi-temporal modelling as one of its key distinctions.

These systems establish that time and provenance belong in the store. They are built primarily for conversational memory: the unit of ingestion is a dialogue turn, the evaluation unit is a question about what was said, and the consumer is a chat loop. Personal agents differ on three counts. Their inputs are artefacts rather than turns, so a single record may describe an event that occurred long before ingestion and may arrive after the event's validity window has already closed. Their identity problem is chronic rather than incidental: the same person is "mom" in one source, a legal name in another, and a kinship term in a third, and resolution may depend on evidence that arrives months later. And their output is not only an answer; it is also a decision about whether to speak at all.

\subsection{Interval semantics and temporal reasoning}

Allen (1983) supplies the interval algebra that any serious temporal store needs; relations such as overlap, during and finish are the primitives for asking whether a fact held during a window. Krause et al. (2022) distinguish validity periods from traceability and argue that the literature conflates dynamic and temporal knowledge graphs, which is the distinction our ingestion-time axis depends on. Tan et al. (2026) show that temporal question answering over knowledge graphs benefits from operator-aware retrieval and a hierarchical decomposition of temporal constraints, which is consistent with treating the temporal focus as a first-class query argument rather than a post-hoc filter.

\subsection{Spatial and spatio-temporal memory}

Embodied agents have produced the most literal spatio-temporal memories. Al Mdfaa et al. (2025) convert egocentric video into a spatiotemporal knowledge graph whose nodes are objects and whose edges are spatial relations over time. Lei et al. (2025) maintain a dynamic knowledge graph as spatial memory for long-horizon planning. Rasheed and Kabtoul (2026) combine graph memory with temporal indexing for embodied agents. Wang et al. (2025) pair long-term scene graphs with short-term object-state memory. Vishwakarma et al. (2025) propose a resonance graph with separate layers for particles, vectors, temporal indexing and relations.

In these systems, space is physical: rooms, objects, metric positions. Personal data has no such coordinate system. A person's life is not indexed by position but by relations: who is whose mother, who paid whom, who accompanied whom to which institution. Our spatial axis is therefore relational rather than metric, We treat the two as complementary: embodied memories index physical space, ours indexes relations.

\subsection{Retrieval}

The retrieval substrate in our prototype follows the standard hybrid recipe. Cormack et al. (2009) show that reciprocal rank fusion of independently ranked lists is competitive with learned rankers and robust without tuning; lexical matching through BM25-style indexes covers names, amounts and identifiers that dense encoders blur, while dense recall covers paraphrase. Wu et al. (2024) provide a long-term memory benchmark that stresses multi-session recall and temporal reasoning, and is, to our knowledge, the closest public instrument to the questions personal agents face. LoCoMo (Maharana et al., 2024) supplies very long multi-session conversations, and Mem0 (Chhikara et al., 2025) describes a production-oriented memory layer; both remain dialogue-centred.

\subsection{Positioning}

Table 1 compares representative systems along the dimensions that matter for personal context. The comparison is qualitative and based on the cited papers' stated designs; it is not a performance comparison.

\begin{table}[t]
\centering
\begingroup\scriptsize
\caption{Qualitative comparison of representative memory systems. \textit{Space} names the unit over which proximity is defined; \textit{scope} names the boundary within which facts are retrieved.}
\begin{tabular}{@{}p{0.97in}p{0.84in}p{0.79in}p{0.58in}p{0.76in}p{0.49in}p{0.84in}@{}}
\toprule
\textbf{System} & \textbf{Time model} & \textbf{Unit of "space"} & \textbf{Provenance} & \textbf{Identity ambiguity} & \textbf{Scope} & \textbf{Beyond question answering} \\
\midrule
MemGPT (Packer et al., 2023) & Memory timestamps & None (tiered text) & Partial & Not modelled & Per agent & No \\
Generative Agents (Park et al., 2023) & Observation time & None (memory stream) & Observation & Not modelled & Single world & No \\
Zep / Graphiti (Rasmussen et al., 2025) & Valid and transaction time & Entity graph & Episodes & Entity nodes & Tenant & No \\
Bitemporal store (Niksarli \& Baheti, 2026) & Valid and transaction time & Versioned graph & Versions & Entity nodes & Per conversation & No \\
SodaMem (Wan et al., 2026) & Mention, occurrence, validity & Fact-event graph & Mandatory spans & Entity nodes & User & No \\
REMem (Shu et al., 2026) & Time-scoped facts and gists & Episodic graph & Gist and fact links & Not central & User & No \\
REAL (Lu et al., 2026) & Validity intervals & Property graph & Assertions & Not central & User & No \\
VL-KnG, STMA, eMEM (Al Mdfaa et al., 2025; Lei et al., 2025; Rasheed \& Kabtoul, 2026) & Frame and episode time & Metric scene graph & Frames & Object identity & Episode & No \\
This framework (proposed) & Event, validity, ingestion, decision window & Relational space plus optional geography & Record pointers, span-level & Candidate sets until evidence resolves & Per data domain & Query, insight, triage, reminder (proposed) \\
\bottomrule
\end{tabular}
\endgroup
\end{table}

Table 1 entries are our reading of each paper's stated design; we did not reproduce any of the systems, and the last row describes a framework whose analytic layer is specified rather than implemented, so it is not compared on the same footing as the deployed systems above.

The closest prior system is SodaMem (Wan et al., 2026), which already requires provenance spans, stores mention time, occurrence time and validity separately, represents supersession and contradiction as typed edges, and indexes both lexically and densely. Three differences remain, and we state them narrowly. First, SodaMem ingests dialogue, whereas our unit is the artefact, which may describe an event that predates ingestion by years. Second, SodaMem resolves mentions to entity nodes, whereas we keep the candidate set as first-class data and let the answering policy carry the ambiguity. Third, SodaMem stops at question answering, whereas we treat silence, triage and interruption as regimes over the same substrate. The gap we target is the intersection of four properties: artefact-first ingestion; unresolved identity preserved as data; answers closed over records; and a consumption model that includes silence. Systems in Table 1 hold two or three of these. We state these four as invariants and make the representation that satisfies them the object of study.

\section{Problem Setting and Definitions}

\subsection{Records}

A record is an immutable observation extracted from one source artefact. Records are the only ground truth in the system; everything else is derived and must be reconstructible from them.

\textbf{Definition 1 (Record).} A record $r$ is a tuple $(id, src, kind, content, t_{evt}, t_{ing}, w, spans)$, where $src$ identifies the source application, $kind$ names the artefact type (message, transfer, photograph, registration slip, itinerary), $content$ is the extracted text or caption, $t_{evt}$ is the event time (a point, an interval, or unknown), $t_{ing}$ is the ingestion time, $w$ is the data domain or scope, and $spans$ locates the supporting region in the source artefact.

The distinction between $t_{evt}$ and $t_{ing}$ is not bookkeeping. A ticket purchased on 26 January for travel on 28 January has an event time in the future and an ingestion time in the present; a photograph taken in 2023 and imported in 2024 has the reverse relation. Queries about what the user knew at a given moment are answerable only if both are retained.

\subsection{Entities, mentions and ambiguity}

\textbf{Definition 2 (Entity and mention).} An entity $e$ is a persistent thing about which the user can ask: a person, an organisation, a place, a project, a pet, an account. A mention $m$ is a surface expression inside a record. The resolution map $C(m) \subseteq E$ returns the candidate set for a mention; $C(m)$ may contain several entities, and it may be revised as records accumulate.

Insisting that $C(m)$ be a singleton at extraction time is a common failure mode in pipelines that resolve at extraction time. In family data, "mom" legitimately denotes two different people: the speaker's mother and the speaker's mother-in-law, both addressed by the same kinship term. Forcing a choice at first sight produces an error that no later evidence can easily undo, because the incorrect link is indistinguishable from a correct one once written. We therefore keep $C(m)$ as a set and attach the evidence that would separate its members. A bank transfer whose memo reads "mom living expenses" and whose payee is recorded under a legal name is evidence about that payee, not a resolution: it raises the posterior that the payee is one of the people the user calls "mom", but a payment to a mother-in-law may carry the same memo. We treat it as separating evidence only in combination with a second source, such as a kinship statement in a document or an explicit user confirmation; until then the candidate set keeps both members.

\subsection{Time}

\textbf{Definition 3 (Interval and validity).} Let $T$ be a linearly ordered time domain. An interval is $[s,e)$ with $s \in T \cup \{-\infty\}$, $e \in T \cup \{+\infty\}$, and $s < e$. For a relation $\rho$, $V(\rho) = [s_\rho, e_\rho)$ is the validity interval: the set of times at which the relation held in the world. Unknown endpoints are represented by open bounds, not by a sentinel date. A point event is written $[t, t+g)$, where $g$ is the finest granule the source supports, because the convention requires $s < e$; one intersection test then serves points and spans alike. Event time and validity are different objects: $t_{evt}$ dates the observation, $V(\rho)$ dates the fact.

We adopt Allen's (1983) relations for interval comparison and use intersection as the fundamental test in retrieval, since most personal queries name a window ("last time", "in January", "during the holiday") rather than a point. An open right endpoint means "still true as far as the system knows", which is a statement about knowledge, not about the world; Section 4.3 makes this explicit in the invariants.

\subsection{Relations}

\textbf{Definition 4 (Relation).} A relation $\rho$ is a tuple $(id, P, \pi, s, V, \sigma, \phi, \omega)$, where $P \subseteq R$ is a non-empty set of supporting records, $\pi \subseteq M$ is the set of participating mentions, $s$ is a natural-language statement, $V$ is the validity interval, $\sigma$ is a status drawn from the state set of Section 4.4, $\phi$ is an optional functional annotation mapping a predicate to a value (for example $\mathrm{location} \mapsto \text{hospital}$, $\mathrm{accompanied} \mapsto \text{self}$), and $\omega$ is the scope of Definition 5.

Relations are the unit of retrieval, and records are the unit of evidence. A relation compresses a record into something queryable; it must never become the thing that is quoted. Section 5.5 turns this into a rule.

\subsection{Scope}

\textbf{Definition 5 (Scope).} A scope $w$ is a named data domain with a boundary the user can reason about: personal, family, one organisation, one project. Records and relations carry their scope, and a query is evaluated within exactly one scope unless the user widens it.

Scope is not a permission system, although it can be enforced as one. It is a semantic boundary: an answer assembled from a work domain and a family domain may be individually correct and jointly misleading, because the two domains use different identity conventions and different notions of what counts as an event. Keeping scope explicit also gives the agent a principled answer to "where did this come from".

\section{The Substrate}

Figure 1 gives the shape of the substrate. Records are the only observations; relations place those observations on two axes; the context operator retrieves only the relations that are simultaneously reachable and valid; and four consumption regimes share the same representation.

\begin{figure}[t]
\centering
\includegraphics[width=\linewidth]{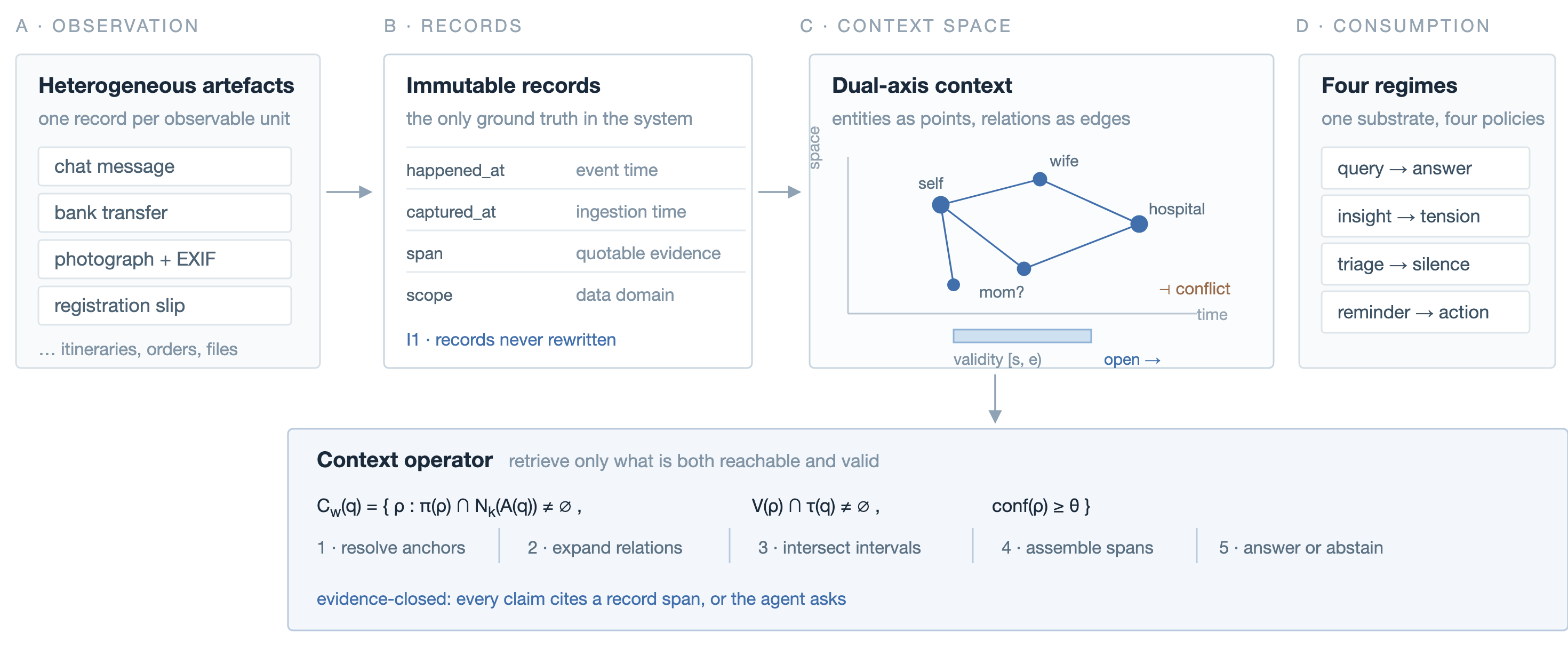}
\caption{The substrate. Source artefacts become immutable records that keep event time and ingestion time apart. Entities and relations form the relational space, with validity intervals on the temporal axis. The context operator composes neighbourhood expansion with interval intersection, and four consumption regimes apply different policies to the same representation.}
\end{figure}

\subsection{The temporal axis}

Three temporal dimensions coexist in the substrate. Event time $t_{evt}(r)$ says when the described event occurred. Validity $V(\rho)$ says over which interval the relation held. Ingestion time $t_{ing}(r)$ says when the system learned it. A fourth, optional dimension is the decision window $d(\rho)$, the deadline by which the user must act on an open relation; it belongs to the consumption layer rather than to the world model, but it is stored with the relation because it is the trigger for proactive behaviour (Section 5.6).

Separating validity from ingestion yields a property that flat memory cannot express: the substrate can answer questions about its own past beliefs. "Did I know about the trip before I paid for it?" is a query over ingestion times; "was the trip still planned on 27 January?" is a query over validity. These are different questions with different answers, and a system that stores a single timestamp will conflate them.

\subsection{The spatial axis}

\textbf{Definition 6 (Context space).} The context space is a graph $G = (E, \mathcal{R})$ whose vertices are entities and whose edges are relations. For an entity $a$, the $k$-neighbourhood $N_k(a)$ is the set of entities reachable within $k$ edges. Proximity may also be measured in an embedding space, but embeddings are an index over statements, not a substitute for the graph.

Two design choices follow. First, distance is relational and typed: a person is near their mother because a kinship edge connects them, not because their home addresses are close. Second, geography is optional. A place entity may carry coordinates, and physical proximity may then be used as an additional signal, but a life is not organised by coordinates and most personal questions never touch them. This is the sense in which our use of "space" differs from the embodied literature: we take the relational structure of meaning as the space, in the way a language model takes a vector space as the place where meaning is arranged.

Figure 2 makes the claim concrete. It shows a short prefix in the same kind of space: twenty-one schematic tokens at fixed positions, the candidates for the next word enlarged and coloured by rank, and an arrow from the current context to the probability-weighted centre of that group. The demonstrator is an explanatory device, not evidence: its positions and probabilities are preset, and nothing in it was produced by a trained model.

\clearpage

\begin{figure}[t]
\centering
\includegraphics[width=\linewidth]{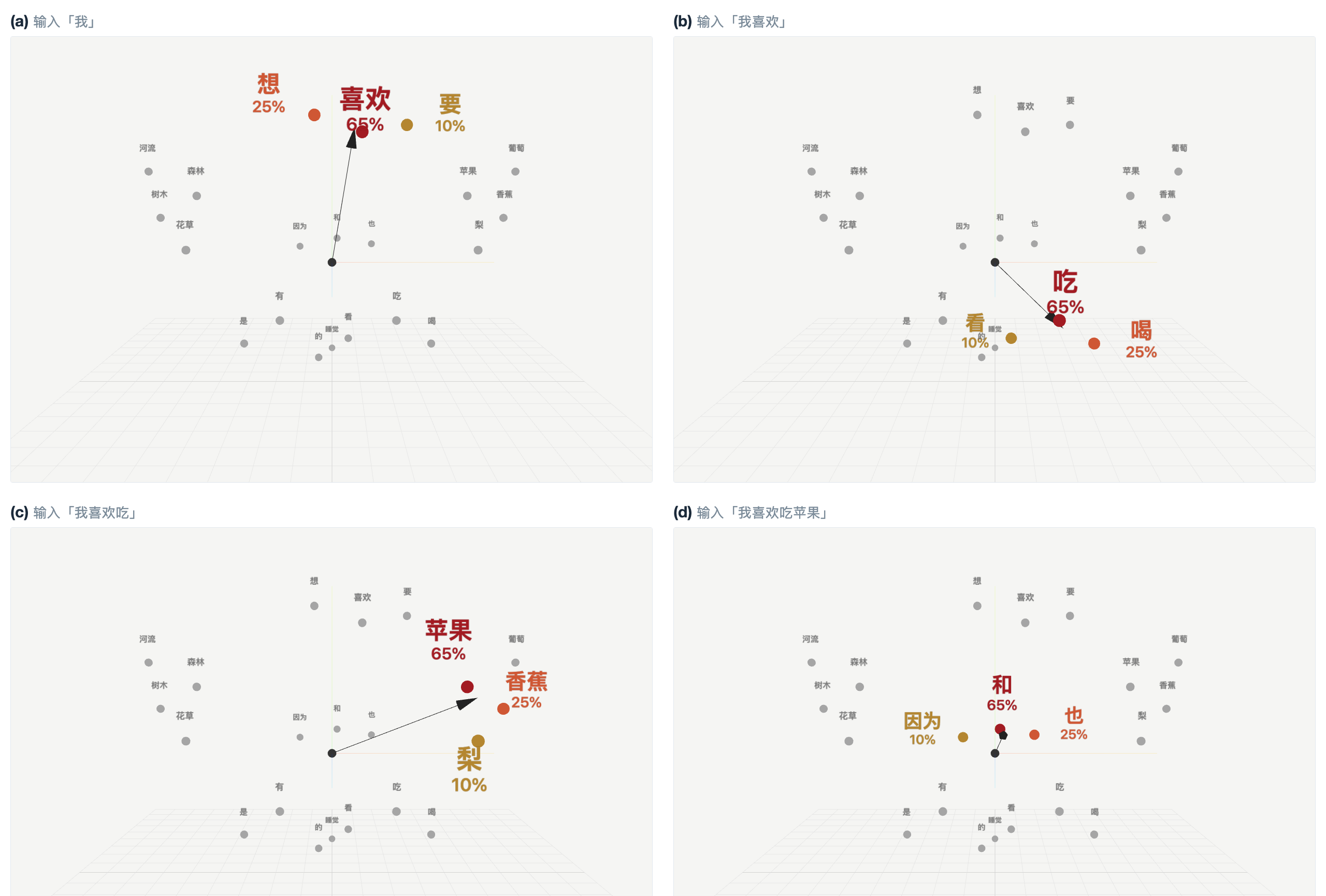}
\caption{A semantic-space demonstrator built for this paper, shown for one Chinese prefix with English glosses in the text. Positions were laid out by semantic neighbourhood, not by projecting a model's embeddings. Each dot is a token at a fixed schematic position, and nearby dots are semantically related. Candidates for the next word are enlarged and coloured by rank with preset probabilities (65\%, 25\%, 10\%); the arrow points to their probability-weighted centre, which lands near, but not on, the top candidate. (a) "I": like 65\%, want 25\%, need 10\%. (b) "I like": eat 65\%, drink 25\%, watch 10\%. (c) "I like eating": apple 65\%, banana 25\%, pear 10\%. (d) "I like eating apples": and 65\%, also 25\%, because 10\%. Positions, probabilities and arrow lengths are schematic: the figure illustrates the geometry of the claim, not the output of a model.}
\end{figure}

\subsection{Invariants}

Six invariants distinguish a context substrate from a memory cache. They are stated as properties the implementation must preserve, and Table 2 pairs each with the failure it prevents.

\begin{table}[t]
\centering
\begingroup\footnotesize
\caption{The six invariants and the failure each prevents.}
\begin{tabular}{@{}p{1.29in}p{2.32in}p{2.32in}@{}}
\toprule
\textbf{Invariant} & \textbf{Statement} & \textbf{Failure prevented} \\
\midrule
I1 Record primacy & Records are immutable and are the only ground truth; entities and relations are derived and rebuildable & Derived errors becoming indistinguishable from observations \\
I2 Multi-time & Event time, validity and ingestion time are stored separately and never collapsed into one timestamp & Confusing when something happened with when it was learned \\
I3 Provenance totality & Every relation cites at least one record; every quoted claim cites a span inside a record & Unsupported assertions and un-auditable answers \\
I4 Open-world validity & An open right endpoint means "still true as far as known", never "current" or "invalid" & Stale facts silently persisting or expiring \\
I5 Ambiguity preservation & A mention may map to a candidate set; resolution requires added evidence, not preference & Premature identity commitments that later evidence cannot repair \\
I6 Scope isolation & Facts do not cross scope boundaries during retrieval unless the user widens the scope & Plausible but incoherent answers assembled across domains \\
\bottomrule
\end{tabular}
\endgroup
\end{table}

The invariants are stated separately because they fail separately. A store that lets an extraction model write a clean triple directly into a graph has traded auditability for convenience; a store that resolves "mom" at ingestion has traded correctness for tidiness.

\subsection{Status and confidence}

Relations move through a small set of states rather than a binary flag. The states are \textit{unresolved} (the participating mention has an open candidate set), \textit{tentative} (extracted but not corroborated), \textit{confirmed} (supported by at least one record and not contradicted), \textit{superseded} (a later relation in the same functional slot continues its validity) and \textit{retracted} (a retraction record has been ingested against it). Withdrawal does not edit the original: a retraction is itself a record, and the retracted state is derived from it, so invariant I1 survives. A superseded relation keeps its records, so history remains queryable. Figure 3 shows the transitions.

\begin{figure}[t]
\centering
\includegraphics[width=\linewidth]{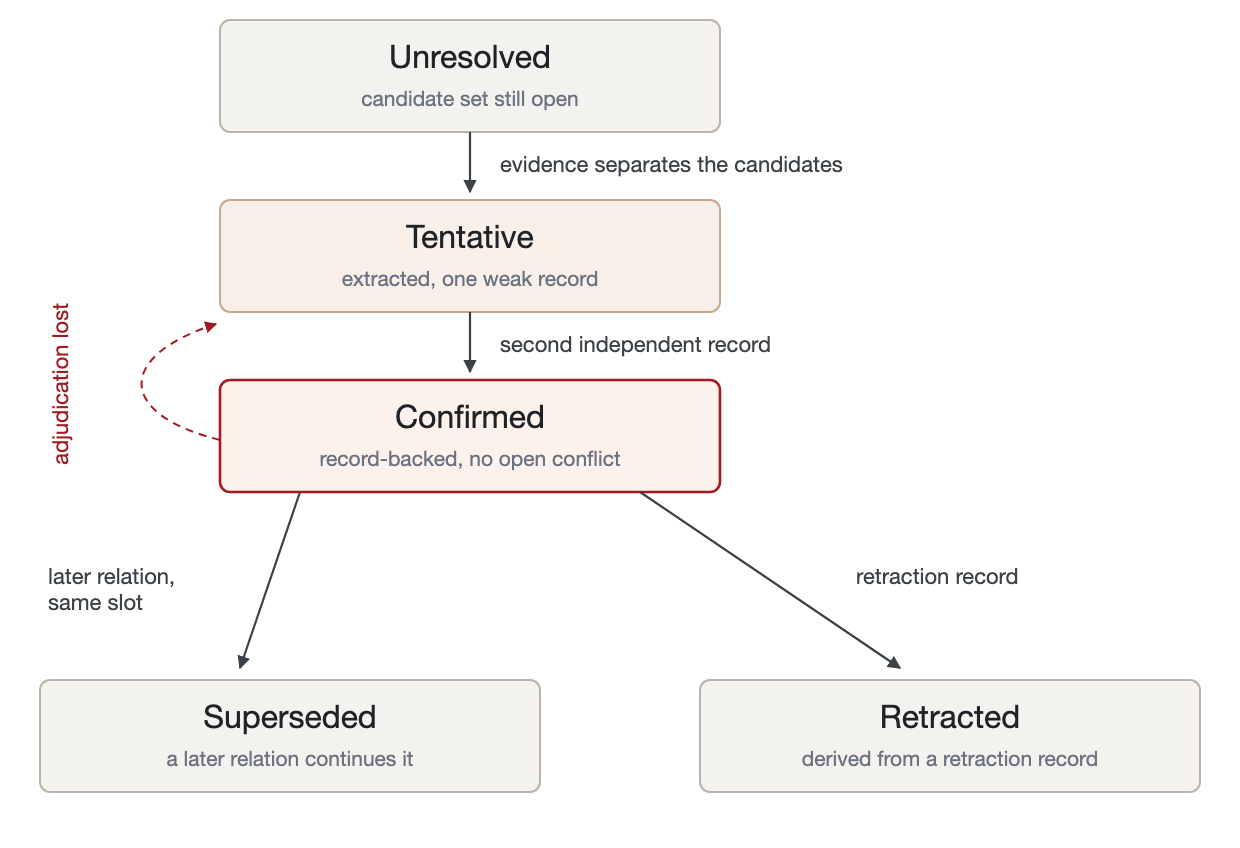}
\caption{Relation status state machine. Edges are labelled with the evidence that licenses them; no transition deletes a record.}
\end{figure}

Transitions are evidence-gated. A relation is created \textit{unresolved} while the participating mention has more than one candidate, and \textit{tentative} when the candidate is fixed but a single weak record supports it. It becomes \textit{confirmed} when provenance exists and no conflict is pending, and it is demoted to tentative when it loses an adjudication. It becomes \textit{superseded} when a later relation in the same functional slot continues its validity, and \textit{retracted} when a retraction record arrives. The trigger layer of Section 5.6 reads status as well as validity, because a conflict that has already been adjudicated should stop generating reminders.

The state machine lets the system be conservative without being silent. A tentative relation can still be retrieved, but it is marked as such in the assembled context, and the answering policy can decide to mention it with a hedge or to withhold it. In the proposed design the status weight is an explicit term in the ranking function, so that a query which can be answered from confirmed relations is not answered from tentative ones. The prototype does not implement that term; Section 7.3 specifies the schema that would carry it.

\section{Operators}

\subsection{The context operator}

A query specifies anchors, a temporal focus and a scope. Anchors are entity mentions or free text that resolvable mentions stand for; the temporal focus is an interval, possibly open; the scope is one domain.

\textbf{Definition 7 (Context).} For a query $q$ with anchor set $A(q)$, temporal focus $\tau(q)$ and scope $w(q)$, and parameters $k$ and $\theta$, the context is

\begin{equation*}
C_w(q) = \{\,\rho \in \mathcal{R}_w \;:\; \pi(\rho) \cap N_k(A(q)) \neq \emptyset,\; V(\rho) \cap \tau(q) \neq \emptyset,\; \mathrm{conf}(\rho) \geq \theta \,\}
\end{equation*}

where $\mathcal{R}_w$ is the set of relations in scope $w$, and $\mathrm{conf}(\rho) \in [0,1]$ is a confidence score that is monotone in the status order of Section 4.4 and in the strength of provenance. The operator composes the two axes: the first condition is spatial expansion in the relational space, the second is temporal intersection, and the third is the confidence floor.

The operator is conjunctive. Disjunction would be a retrieval convenience, but it would admit relations that are temporally right and relationally irrelevant, or vice versa, which is the failure mode we design against. The parameters expose the usual precision-recall tradeoff: larger $k$ widens the neighbourhood, a wider $\tau$ widens the window, and a lower $\theta$ admits weaker evidence.

\subsection{Anchor resolution}

Resolving an anchor proceeds in two channels. The lexical channel matches surface forms and aliases exactly, which is how names, amounts and identifiers are found. The semantic channel embeds the query and retrieves the top-$k$ entity summaries and relation statements, which is how paraphrases and unnamed references are found. The two channels are fused by reciprocal rank fusion (Cormack et al., 2009), which requires no score calibration and is therefore appropriate when lexical and dense scores live on incomparable scales.

The output of anchor resolution is not a single entity but a candidate set with supporting evidence. If the set is a singleton, the anchor is resolved. If it has several members, the anchor is ambiguous, and the ambiguity propagates into the answering policy rather than being resolved by a preference for frequency or recency.

\textbf{Answer-consistency test.} Let $C(m) = \{e_1, \dots, e_j\}$ be the candidate set for a mention that the query depends on. For each candidate, assemble the context $C_w(q \mid e_i)$ and derive an answer $a_i$. If all $a_i$ agree on the queried predicate, return that answer and record the resolution that produced it. If they disagree, return a disambiguation request that names the candidates and the evidence that separates them, rather than choosing one. This rule converts an unavoidable uncertainty into a cheap question instead of a confident error, and it is the operational meaning of invariant I5. It also has a cost that grows quickly: with ambiguity on $j$ mentions whose candidate sets have sizes $m_1,\dots,m_j$, the test assembles $\prod_i m_i$ contexts. The prototype therefore caps $j$ at two and expands only mentions that can change the answer; Section 10 records this as a limitation.

\subsection{Temporal filtering}

Temporal filtering uses interval intersection rather than date equality. Given $V(\rho) = [s,e)$ and $\tau(q) = [s_\tau, e_\tau)$, the relation is in scope when $s < e_\tau$ and $s_\tau < e$. The comparison is cheap, but it must be applied to the right object: filtering relations rather than records is what makes "my mother was in hospital in January" answerable when the registration slip carries an event time inside January, independently of whether the accompanying relation is still open.

Three practical cases recur. A point event is a one-granule interval (Section 3.3), so a single comparison handles both points and spans. An unknown endpoint is open, so a fact learned once does not silently expire. A query with an unknown focus ("last time") is resolved by ordering candidate intervals and taking the most recent confirmed one, with ties broken by ingestion time. Ingestion time is used only as a tie-breaker, never as evidence of correction. When two records claim the same event time and are mutually exclusive, the store prefers the relation with stronger provenance and, failing that, asks the user; a later ingestion date means the system learned the fact later, not that the fact is truer.

\subsection{Conflict detection and adjudication}

Personal data contains two kinds of tension that look alike in a similarity index: records that are wrong, and records that are right but mutually exclusive. A store that averages them can produce a statement that no single record supports. A store that overwrites the older fact produces a silent loss of history. We separate them by making functional exclusivity explicit.

\textbf{Definition 8 (Functional conflict).} Let $\mathcal{F}$ be a set of functional constraints, each a pair $(f, \bot)$ where $f$ maps a relation to a value and $\bot$ is an exclusivity relation on that value domain. Two relations conflict when

\begin{equation*}
V(\rho_1) \cap V(\rho_2) \neq \emptyset \;\wedge\; \exists (f,\bot) \in \mathcal{F} : f(\rho_1) \,\bot\, f(\rho_2).
\end{equation*}

Typical constraints are $\mathrm{location}(\text{person}) \mapsto \text{place}$ with exclusivity "at most one place at a time", and $\mathrm{employment}(\text{person}) \mapsto \text{organisation}$ with exclusivity "at most one active employer". The first is what turns two individually plausible records into a question for the user: a person cannot be accompanying a relative in one city and travelling to another at the same moment. The constraint is evaluated on validity intervals, not on calendar days: when the evidence fixes only dates, the pair is reported as a \textit{potential} conflict and left to the user, because a person can accompany a relative in the morning and travel in the evening.

The adjudication policy is a three-way decision. If both relations are confirmed and their records are strong, the system surfaces the tension as a decision request and does not choose. If one relation is tentative or its record is weak, the system demotes it and marks the pair as unverified, which keeps the context usable without bothering the user. If the conflicting relations have disjoint validity or belong to different functional slots, no action is taken. The thresholds are a precision-recall knob over interruptions. Their calibration is left to the implementation, and Section 9 makes interruption cost a measured quantity rather than an assumption.

Adjudication is itself recorded. When the user resolves a tension, the decision is stored as a new record with its own ingestion time and scope, so that the resolution has provenance like any other fact. This is the mechanism by which a personal agent's context improves without retraining: we expect user decisions to be among the highest-confidence records in the store, although this is an assumption we do not test here.

\subsection{Evidence-closed answering}

The assembly step produces a set of relations with record pointers. Generation must be closed over those records.

\textbf{Definition 9 (Evidence closure).} An answer $A$ decomposes into atomic claims $c_1, \dots, c_n$. Let $\mathrm{supp}(c_i)$ be the set of record spans cited for $c_i$. The answer is evidence-closed when every factual claim has a non-empty support whose spans jointly entail the claim. Citation coverage is $\kappa(A) = |\{i : \mathrm{supp}(c_i) \neq \emptyset\}| / n$, and evidence closure requires $\kappa(A) = 1$ for factual claims.

Intermediate relations are not citable. The chain that connects a chat message to a bank transfer to a hospital visit is a retrieval path, not evidence; the records the path traversed are, and all of them must be listed. Joint entailment matters precisely for such derived claims: the kinship in Section 8 is supported by a registration slip, a transfer and a chat message together, and no single span entails it. When closure fails the agent must abstain or ask, not paraphrase. Abstention is the intended behaviour here, and the alternative --- a plausible sentence whose support is a similarity score --- is the failure the rule exists to prevent.

The rule also disciplines the prompt. The generator receives relations for orientation and records for citation, and it is instructed to attach a record identifier and a span to each claim. A verifier checks that the cited span exists and that the claim does not assert more than the span supports. This division of labour is what SodaMem (Wan et al., 2026) formalises with mandatory provenance spans, and we adopt it as an invariant rather than an optional feature. The verifier is part of the specified analytic layer, not of the prototype; whether it is an entailment model, a prompted model or a rule set is left open, because any such verifier has an error rate and Section 9 treats verifier error as a threat to validity rather than as a guarantee.

\subsection{Trigger evaluation}

Proactive behaviour is a decision, not a side effect of retrieval. A relation is \textit{open} when its right endpoint is unbounded and a state predicate is unfinished, such as an unpaid ticket or an unresolved conflict.

\textbf{Definition 10 (Decision window).} For an open relation $\rho$ with deadline $d(\rho)$, the decision window at time $t$ is the interval $[d(\rho) - \theta, d(\rho)]$. A trigger fires when $t$ lies in the window, the relation remains open, and the expected value of acting exceeds the cost of interrupting.

The value and cost terms are left unspecified here; any concrete policy must set them. The point of the definition is that interruption is gated by three conditions that must all hold, rather than by a single relevance score. Two failure modes bound the design: a reminder that fires on relevance alone becomes a notification stream, and a reminder that never fires because the evidence is incomplete is silent. The three-condition gate is meant to sit between them; Section 9 measures the first failure directly as interruption rejection, whereas the second is visible only through the abstention metric. A concrete policy would estimate the value of acting from the consequences of missing the deadline, and the cost of interrupting from the user's recent responsiveness and the channel. Section 9 treats false interruptions as a first-class error.

\subsection{Pipeline and complexity}

Table 3 states the pipeline and marks which stages the prototype implements. Exact matching and vector recall run in parallel and are fused before any filtering; the temporal and relational filters and the evidence assembly belong to the specified analytic layer.

\begin{table}[t]
\centering
\begingroup\footnotesize
\caption{Retrieval pipeline stages, cost profile and implementation status.}
\begin{tabular}{@{}p{0.77in}p{1.53in}p{1.01in}p{0.93in}p{1.34in}@{}}
\toprule
\textbf{Stage} & \textbf{Operation} & \textbf{Output} & \textbf{Cost note} & \textbf{Status} \\
\midrule
1 Decompose & Split the query into anchors, temporal focus and scope & Query frame & Linear in query length & Implemented, by heuristics \\
2 Anchor resolution & Alias match plus dense top-$k$, fused by reciprocal rank fusion & Candidate entities and mentions & Two index lookups & Implemented \\
3 Relational expansion & $k$-hop neighbourhood over entity and relation edges & Candidate relation set & Bounded by degree and $k$ & Proposed \\
4 Temporal filter & Interval intersection with the focus & Time-valid relation set & Interval index & Proposed \\
5 Evidence assembly & Replace relations by their record pointers and spans & Evidence bundle & Linear in selected relations & Proposed \\
6 Answer & Generate with records only, verify spans, abstain on failure & Evidence-closed answer & Model-bound & Generation implemented, closure not verified \\
\bottomrule
\end{tabular}
\endgroup
\end{table}

The dominant cost is generation, which is why stage 5 is a compression step: the bundle passed to the model should contain the spans that support the answer rather than the neighbourhood that produced them. On the deployed store the two recall channels differ by four orders of magnitude. A BM25 top-50 lookup returns in 0.07 ms at the median and 0.44 ms at p95 over 24 queries; the dense channel, which scores every stored vector by cosine similarity in process, costs 0.85 s at the median. The dense scan therefore dominates retrieval and grows linearly with the corpus, which is the price of avoiding a vector database. At the measured scale it is tolerable in an interactive client; at ten times the scale an approximate index becomes necessary.

\section{Consumption Regimes}

A context substrate is used in four ways, and each imposes a different requirement on the representation. Table 4 summarises them; the prose that follows explains why the distinctions matter.

\begin{table}[t]
\centering
\begingroup\footnotesize
\caption{Four consumption regimes over one substrate.}
\begin{tabular}{@{}p{0.52in}p{1.23in}p{1.51in}p{1.15in}p{1.18in}@{}}
\toprule
\textbf{Regime} & \textbf{Question shape} & \textbf{Operator composition} & \textbf{Output} & \textbf{Failure guarded} \\
\midrule
Query & "When, where, with whom" & Anchor resolution, temporal filter, evidence assembly & Evidence-closed answer & Confident answers without support \\
Insight & "Is anything inconsistent or changing" & Relational expansion plus conflict detection over a window & Tension report or trend & Silent averaging of incompatible facts \\
Triage & "Does this deserve the user's attention" & Scoped lookup plus policy evaluation & Silent handling, reply, or escalation & Notification flooding \\
Reminder & "What must be decided before a deadline" & Open-relation scan plus decision window & Proactive message with actions & Deadlines discovered too late \\
\bottomrule
\end{tabular}
\endgroup
\end{table}

Query is the regime most studied in prior work: a question is given, evidence is assembled, an answer is returned. Insight is harder because nothing is asked; the system must compare relations that were never designed to be compared, which is exactly the functional-conflict machinery of Section 5.4. A conflict review over a two-week window is a query whose predicate is "any pair of confirmed relations that violate a functional constraint", and its output is a question for the user rather than an answer.

Triage handles events the user never asked about. Incoming events are often not for the user: a marketing call, a recruiter's message, a distant acquaintance asking a question whose answer is already in the store. Handling these silently requires the same context as answering them, plus a policy layer that decides whether the event touches an entity the user cares about and whether the answer is already known. The value of the substrate here is that "does this concern me" becomes a lookup over relations rather than a judgement over text.

Reminder is the regime that makes the substrate proactive. Its trigger is the decision window of Section 5.6, and its output must include the evidence that justified the interruption, because an unexplained reminder is indistinguishable from a notification. The strongest reminders cite two or three records whose combination creates the obligation: a ticket that is unpaid, a date that is imminent, and an open relation that makes travel impractical.

The four regimes share one substrate and differ in policy. This is the argument for a substrate rather than four features: the expensive part is the representation, and we expect the marginal cost of a further regime to be low once the operators exist.

\section{Implementation}

\subsection{Local-first store}

The prototype is a desktop agent whose knowledge base lives on the user's machine. The retrieval layer of Sections 7.1 and 7.2 is deployed in an internal build. The source is not public: this paper releases no code, no dataset and no supplementary file. The analytic layer of Section 7.3 is specified, not implemented. The store is SQLite in write-ahead logging mode with foreign keys enforced, which gives transactional ingestion, crash safety and zero operational dependency. Ingestion is incremental: files are identified by content hash, parsed in a worker process, chunked by structure, embedded, and indexed, with a status machine that survives restarts and marks partial work explicitly.

Chunking is structure-aware rather than fixed-width. Documents are split along heading boundaries and page markers, paragraphs are merged to a target of roughly 400 tokens, adjacent chunks keep an overlap of roughly 80 tokens, and code blocks are kept intact. Images produce two chunks, a textual chunk carrying OCR output and a visual chunk carrying the image itself, so that a caption and its picture remain separately retrievable and jointly citable.

\subsection{Hybrid recall}

Recall runs two channels in parallel. The lexical channel uses an inverted full-text index with BM25 scoring and a Chinese word segmenter applied before indexing, which is necessary because whitespace tokenisation does not segment Chinese text. The dense channel embeds queries and passages with a 768-dimensional on-device model (EmbeddingGemma 2, INT8, executed through ONNX Runtime), so that no text leaves the machine. The two ranked lists are fused with reciprocal rank fusion at $k = 60$, recalling the top 50 candidates per channel and passing the top 10 to the caller by default. Vectors are stored as float32 blobs and scored by cosine similarity in-process, which removes the need for a vector database and fixes retrieval cost at linear in the number of vectors.

Multimodal material is mapped into the same index. Video contributes time-windowed visual chunks and audio-window chunks, images contribute visual and textual chunks, and EXIF metadata contributes capture time and optional coordinates as entity attributes. This is what allows a photograph to be treated as a record with an event time rather than as an opaque binary.

\textbf{Measured cost.} On the deployed store, the corpus holds 341 documents, 340,079 chunks and 157,440,727 tokens in a 3.46 GB SQLite file; 190,271 chunks carry a 768-dimensional vector, 584 MB of float32. Over 24 queries with a warm cache, the lexical channel's top-50 lookup takes 0.07 ms at the median and 0.44 ms at p95 (maximum 0.64 ms). The dense channel scans all vectors in process and takes 0.85 s at the median (range 0.83--0.89 s). These figures come from a harness that reads the store directly rather than from instrumenting the client, so they exclude query embedding, fusion policy beyond the two lookups, and generation. They were taken on a single Apple-silicon desktop with the database on local SSD, and they are not a comparison against other systems.

\subsection{The analytic layer}

The retrieval substrate described above is implemented in the desktop client. The entity-and-relation layer that this paper argues for is specified here as the next layer, and Table 5 gives its schema as the concrete form of Definitions 1, 2 and 4.

\begin{table}[t]
\centering
\begingroup\footnotesize
\caption{Proposed analytic schema for the entity-and-relation layer.}
\begin{tabular}{@{}p{0.97in}p{2.48in}p{2.48in}@{}}
\toprule
\textbf{Table} & \textbf{Fields} & \textbf{Role} \\
\midrule
\texttt{records} & id, source\_id, kind, content, happened\_at, captured\_at, scope, span & Immutable observations; adjudications and retractions are records of their own kind \\
\texttt{entities} & id, type, name, aliases, first\_seen, embedding, status & Persistent things; candidate sets live in an association table keyed by mention \\
\texttt{mentions} & mention\_id, record\_id, surface, candidate\_entity\_id, evidence\_record\_id & Ambiguity as data rather than as a decision \\
\texttt{relations} & id, statement, participants, valid\_from, valid\_to, scope, decision\_deadline, status, functional\_key, embedding & Queryable facts with provenance pointers, validity, scope and the decision window \\
\texttt{relation\_records} & relation\_id, record\_id, span, weight & The supporting set of each relation; invariant I3 is enforced here \\
\texttt{adjudications} & relation\_a, relation\_b, decision, decided\_at, note, record\_id & User decisions as records with provenance \\
\bottomrule
\end{tabular}
\endgroup
\end{table}

Three schema choices deserve comment. Relations point to records through \texttt{relation\_records} rather than embedding statements as their source of truth, which enforces invariant I3 at the storage level and makes a supporting set, rather than a single pointer, the unit of evidence. Validity is stored as two nullable bounds rather than as a single timestamp, so that open intervals are representable and queryable, which enforces I4. And the decision window, which belongs to the consumption layer, is stored with the relation rather than recomputed, because the trigger rule reads it on every scan.

\subsection{Scope and privacy}

Because the store is local, scope boundaries are physical as well as logical: in the prototype each library is a separate database file, and retrieval cannot cross a library boundary by accident. This matters more than it might appear. The alternative, a single store with logical scopes, relies on policy for a property that a per-library file gives structurally, and we consider structural isolation the simpler guarantee to audit.

\section{Illustrative Walkthrough}

The following vignette is constructed for exposition; the records are invented and the numbers are not measurements. Its purpose is to show how the operators compose, not to report performance.

Five records arrive from five sources. A family chat message from November reads "Are you coming home for the New Year? The cured meat is ready." A bank transfer in December records a payment to a legal name with the memo "mom living expenses". A wedding photograph imported later carries a handwritten note on its reverse naming a mother-in-law. A hospital registration slip lists a patient, an accompanying person and the date 12 January. A travel itinerary, ingested on 26 January, records an unpaid ticket for 28 January.

\begin{table}[t]
\centering
\begingroup\footnotesize
\caption{The operators traced over the constructed vignette.}
\begin{tabular}{@{}p{1.04in}p{2.44in}p{2.44in}@{}}
\toprule
\textbf{Stage} & \textbf{Operation on the vignette} & \textbf{Result} \\
\midrule
Anchor resolution & "mom" yields two candidates: the speaker's mother and the mother-in-law named on the photograph & Candidate set of size two; the transfer memo raises the posterior for one member but does not separate them \\
Background & The registration slip lists the patient and the speaker as accompanying person & A kinship statement that names the patient as the speaker's mother, which separates the candidate set \\
Temporal filter & Focus is January & The slip survives as a one-granule interval on 12 January; the ticket is a point event on 28 January and falls outside \\
Conflict detection & Location is functional and exclusive & The accompaniment on 12 January and the travel on 28 January are sixteen days apart, so no conflict is raised \\
Evidence assembly & Relations are replaced by their records & Slip, transfer and chat message, each with its span \\
Answer & The question asks where the speaker was on 27 January & The answer states the 12 January facts and abstains on 27 January, citing the three records that support the kinship \\
Insight & "Anything inconsistent in the next two weeks?" & The unpaid ticket and the absence of a return record make a \textit{potential} tension, which is surfaced as a question \\
Reminder & 27 January, ticket unpaid, travel one day away & Fires, naming the itinerary record and the amount due \\
Triage & A call from an unknown number & No relation connects the caller to any entity the user cares about; the call is declined and does not ring \\
\bottomrule
\end{tabular}
\endgroup
\end{table}

The query "when my mother was last in hospital, where was I?" resolves through the slip and the transfer. The accompaniment relation is valid on 12 January, and the answer reports that the speaker is listed as the accompanying person on that date. It does not extend the relation to 27 January, because an open right endpoint is a statement about what the store knows, not about where the speaker was. The kinship chain that connects the message, the transfer and the slip is a retrieval path; each record it traversed is cited, and the derived kinship itself is stated as derived.

The insight query no longer returns the 28 January conflict, because the temporal filter has moved the travel outside the January window. What it returns instead is weaker: an unpaid ticket with no return record and no cancellation, which is a potential tension rather than a contradiction, and the agent asks rather than asserts. That asymmetry between a clean contradiction and a suggestive gap is the honest output of a substrate that refuses to average.

The reminder fires on 27 January for the three stated reasons: the ticket is unpaid, the travel date is one day away, and the accompanying relation from 12 January gives the trip a purpose it may no longer have. The message names the records and offers two actions, because an unexplained reminder is indistinguishable from a notification. The triage case shows the same substrate staying silent: the caller has no relation to any entity the user cares about, so the correct output is no output.

\section{Evaluation Protocol}

The framework needs an evaluation that measures what the invariants promise. We specify four research questions and the instruments for each. None of the protocol has been run: the only measurements in this paper are the retrieval timings of Section 7.2. We state the protocol so that the remaining claims can be tested rather than believed.

RQ1 asks whether evidence-closed answering reduces unsupported claims without unacceptable abstention. The instrument is citation coverage $\kappa$ paired with an entailment check on cited spans, plus an abstention rate. A system that achieves closure by refusing everything is caught by the abstention term. Because the entailment check is itself a model, RQ1 also reports the verifier's accuracy on a held-out set of span-claim pairs, so that closure is not confused with verifier agreement.

RQ2 asks whether the dual-axis operator improves temporal questions relative to similarity retrieval. The instrument is accuracy on time-scoped questions, stratified by whether the answer requires interval intersection, entity resolution, or both. LongMemEval (Wu et al., 2024) supplies multi-session and temporal categories; Zep (Rasmussen et al., 2025) reports on a conversational benchmark of its own, which we cite as a comparison point rather than adopt as an instrument. LongMemEval is dialogue data, so it exercises the temporal axis but not ambiguity preservation. Neither covers artefact-shaped personal data, which motivates a constructed set.

RQ3 asks whether conflict detection separates world-level tension from data error. The instrument is precision and recall over adjudications, with a human label for each pair as genuine tension, erroneous record, or non-conflict. This is the metric most likely to be gamed by a detector that flags everything, so precision at a fixed interruption budget is the primary number.

RQ4 asks what the substrate costs the user in attention. The instrument is the rate of proactive messages that the user rejects, plus the latency and token cost of answering.

Table 7 lists the metrics, their definitions and the instrument for each.

\begin{table}[t]
\centering
\begingroup\footnotesize
\caption{Evaluation metrics and instruments.}
\begin{tabular}{@{}p{1.46in}p{2.33in}p{2.14in}@{}}
\toprule
\textbf{Metric} & \textbf{Definition} & \textbf{Primary instrument} \\
\midrule
Citation coverage $\kappa$ & Fraction of factual claims with an entailing span & Constructed claim-level annotations \\
Verifier accuracy & Agreement of the span-claim entailment check with human labels & Held-out span-claim pairs \\
Abstention rate & Fraction of answerable queries answered with a refusal & Annotated query set \\
Temporal accuracy & Correctness on queries requiring interval reasoning & LongMemEval temporal split plus constructed set \\
Resolution accuracy & Correct entity chosen when evidence separates candidates & Constructed ambiguity set \\
Conflict precision at budget & Precision of adjudications at a fixed rate of interruptions & Human labels on conflicting pairs \\
Interruption rejection rate & Proactive messages the user declines or ignores & Interaction logs \\
Reconstruction agreement (I1) & Fraction of stored relations that can be rebuilt from their records and match & Rebuild harness over a frontier snapshot \\
Scope leakage rate (I6) & Answers that draw on facts outside the requested scope & Constructed cross-domain queries \\
Cost & Latency and tokens per answered query & Instrumented runs \\
\bottomrule
\end{tabular}
\endgroup
\end{table}

Ablations follow directly from the invariants. Removing the temporal filter tests I2 and I4. Removing provenance and allowing relation-level citation tests I3. Forcing resolution at ingestion tests I5. Merging scopes tests I6. We expect each ablation to degrade a specific metric rather than the aggregate, which is the point of stating invariants separately.

Five threats are structural. First, personal data is scarce and sensitive, so a public benchmark cannot simply be collected, and constructed suites risk measuring the constructor's imagination. Second, the ground truth for identity and conflict is the user's own judgement, which is stable but expensive, and disagreements between annotators may reflect genuine ambiguity rather than annotation error. Third, extraction and entailment are performed by models, so a closed answer is only as correct as the extractor that produced its records and the verifier that approved its spans; both error rates have to be reported alongside the answer metrics. Fourth, the prototype is single-user, so nothing here speaks to shared or partially visible contexts. Fifth, the vignette is constructed, and no user study has been run, which means the interruption metrics of RQ4 are specified but not observed. These threats argue for a protocol that reports per-invariant metrics on small, well-annotated sets rather than a single aggregate score on a large, weakly labelled one.

\section{Discussion and Limitations}

The framework's central bet is that explicitness beats capacity. A larger model with a larger context window can hold more text, but it cannot tell which sentence was learned when, which two sentences describe the same person, or which sentence licenses which claim. Those are properties of a representation, and a longer context alone does not supply them. The bet could be wrong in a specific way: if extraction is sufficiently reliable and models are sufficiently good at implicit coreference and temporal reasoning, an explicit substrate may be unnecessary overhead. We think the opposite is more likely for personal data, where sources are heterogeneous and mutually inconsistent in naming and convention, but this is a claim to be tested rather than asserted.

The costs are real. Extraction errors propagate: a wrong relation is worse than a missing one, because it is retrievable and looks authoritative. Our invariants reduce the damage rather than eliminate it, since a relation that cites a real record can still misread that record. Interval reasoning adds bookkeeping that most queries never need. Validity is cheap to store, whereas status and conflict are expensive to maintain, so the substrate derives the latter lazily, for the entities and windows a query touches, rather than eagerly for the whole corpus. Ambiguity preservation pushes work into the answering policy, where it becomes a user-visible question; a user who is asked to disambiguate too often will experience the agent as pedantic, and the calibration of when to ask is unsolved. The consistency test also multiplies: with $j$ ambiguous mentions, the number of contexts to assemble is the product of the candidate-set sizes, which is why the prototype caps the expansion at two mentions. A cheaper version would rank candidates by the expected change in the answer, and we have not implemented one.

Local-first computation bounds what is possible. On-device embeddings at 768 dimensions and in-process cosine scoring buy privacy that a cloud index cannot offer, but the measured cost of the scan (Section 7.2) is linear in the number of vectors, and they rule out the largest rerankers and the most expensive extraction models. The design therefore assumes a division of labour: local retrieval and local embedding for recall, a hosted or local large model for generation, and no raw content leaving the machine unless the user chooses.

Scope and multi-user settings are underexplored. A family shares entities but not all facts; an organisation shares facts but not all entities. Our scope is a flat domain, which is enough for a single user with a few domains and insufficient for shared contexts with partial visibility. The natural extension is scope as a partial order with explicit joins, which will change the retrieval operator rather than merely the permission check.

Finally, the framework inherits the evaluation problem of personal agents. Aggregate accuracy hides the failures that matter, which are the confident wrong answer, the missed deadline and the unwanted interruption. The metrics in Section 9 are an attempt to make those failures visible; they are not a solution to the problem of obtaining labelled personal data. What this paper can claim at present is narrower than the framework's ambition: the retrieval layer exists and has been measured, and the analytic layer that the invariants describe is a design with a stated test plan.

\section{Conclusion}

The argument of this paper is that what limits personal agents is the representation of the user's world rather than the size of the model. We have argued that this representation has two axes, time and a relational space, and that it must satisfy six invariants: records as ground truth, multi-time, provenance totality, open-world validity, ambiguity preservation and scope isolation. From those invariants we derived a context operator that composes temporal intersection with relational expansion, an answer-consistency test that turns unresolved identity into a question rather than an error, a functional-exclusivity detector that converts incompatible facts into adjudication rather than overwrite, an evidence-closure rule that forbids citing derivations, and a trigger rule that gates interruption on open relations and deadlines. We classified four consumption regimes over one substrate, described an implemented retrieval layer, specified the analytic layer that the invariants require, and set out an evaluation protocol with per-invariant metrics.

The broader claim is that the unit of design for agent context is not the memory but the substrate: a set of invariants that make the store auditable, a pair of axes that make it queryable, and a policy layer that decides when the agent should speak. The next steps are to implement the analytic layer, run the protocol of Section 9 on consented personal data, and extend scope from a flat domain to a partial order over shared contexts.

\section*{References}
\begingroup\small

\par\noindent\hangindent=1.2em\hangafter=1 Allen, J. F. (1983). Maintaining knowledge about temporal intervals. \textit{Communications of the ACM, 26}(11), 832--843. \url{https://doi.org/10.1145/182.358434}\par\vspace{1pt}

\par\noindent\hangindent=1.2em\hangafter=1 Al Mdfaa, M., Lukina, S., Akhtyamov, T., Nigmatzyanov, A., Nalberskii, D., Zagoruyko, S., \& Ferrer, G. (2025). Spatiotemporal knowledge graphs as persistent scene memory for embodied question answering (VL-KnG). arXiv:2510.01483v3. \url{https://arxiv.org/abs/2510.01483}\par\vspace{1pt}

\par\noindent\hangindent=1.2em\hangafter=1 Balog, K., \& Kenter, T. (2019). Personal knowledge graphs: A research agenda. In \textit{Proceedings of the 2019 ACM SIGIR International Conference on Theory of Information Retrieval} (pp. 217--220). \url{https://doi.org/10.1145/3341981.3344241}\par\vspace{1pt}

\par\noindent\hangindent=1.2em\hangafter=1 Chhikara, P., Khant, D., Aryan, S., Singh, T., \& Yadav, D. (2025). Mem0: Building production-ready AI agents with scalable long-term memory. arXiv:2504.19413. \url{https://arxiv.org/abs/2504.19413}\par\vspace{1pt}

\par\noindent\hangindent=1.2em\hangafter=1 Cormack, G. V., Clarke, C. L. A., \& B\"{u}ttcher, S. (2009). Reciprocal rank fusion outperforms Condorcet and individual rank learning methods. In \textit{Proceedings of the 32nd International ACM SIGIR Conference} (pp. 758--759). \url{https://doi.org/10.1145/1571941.1572114}\par\vspace{1pt}

\par\noindent\hangindent=1.2em\hangafter=1 Krause, F., Weller, T., \& Paulheim, H. (2022). On a generalized framework for time-aware knowledge graphs. In \textit{Semantic Web Challenges} (Studies on the Semantic Web). \url{https://doi.org/10.3233/SSW220010}\par\vspace{1pt}

\par\noindent\hangindent=1.2em\hangafter=1 Lei, M., Zhao, Y., Wang, G., Mai, Z., Cui, S. R., \& Han, Y. (2025). STMA: A spatio-temporal memory agent for long-horizon embodied task planning. arXiv:2502.10177. \url{https://arxiv.org/abs/2502.10177}\par\vspace{1pt}

\par\noindent\hangindent=1.2em\hangafter=1 Lu, K., Chen, L., Jiang, G., Qin, Z., Liu, Y., \& Zhang, W. (2026). REAL: A reasoning-enhanced graph framework for long-term memory management of LLMs. arXiv:2606.10694. \url{https://arxiv.org/abs/2606.10694}\par\vspace{1pt}

\par\noindent\hangindent=1.2em\hangafter=1 Lupa\c{s}cu, C., \& Lupascu, A. (2026). ElephantBroker: A knowledge-grounded cognitive runtime for trustworthy AI agents. arXiv:2603.25097. \url{https://arxiv.org/abs/2603.25097}\par\vspace{1pt}

\par\noindent\hangindent=1.2em\hangafter=1 Maharana, A., Lee, D.-H., Tulyakov, S., Bansal, M., Barbieri, F., \& Fang, Y. (2024). Evaluating very long-term conversational memory of LLM agents. arXiv:2402.17753. \url{https://arxiv.org/abs/2402.17753}\par\vspace{1pt}

\par\noindent\hangindent=1.2em\hangafter=1 Niksarli, A., \& Baheti, G. (2026). A graph-native bitemporal memory store for conversational AI agents. arXiv:2607.26520. \url{https://arxiv.org/abs/2607.26520}\par\vspace{1pt}

\par\noindent\hangindent=1.2em\hangafter=1 Packer, C., Wooders, S., Lin, K., Fang, V., Patil, S. G., \& Stoica, I. (2023). MemGPT: Towards LLMs as operating systems. arXiv:2310.08560. \url{https://arxiv.org/abs/2310.08560}\par\vspace{1pt}

\par\noindent\hangindent=1.2em\hangafter=1 Park, J. S., O'Brien, J., Cai, C. J., Morris, M. R., Liang, P., \& Bernstein, M. S. (2023). Generative agents: Interactive simulacra of human behavior. In \textit{Proceedings of the 36th Annual ACM Symposium on User Interface Software and Technology}. \url{https://doi.org/10.1145/3586183.3606763}\par\vspace{1pt}

\par\noindent\hangindent=1.2em\hangafter=1 Rasmussen, P., Paliychuk, P., Beauvais, T., Ryan, J., \& Chalef, D. (2025). Zep: A temporal knowledge graph architecture for agent memory. arXiv:2501.13956. \url{https://arxiv.org/abs/2501.13956}\par\vspace{1pt}

\par\noindent\hangindent=1.2em\hangafter=1 Rasheed, A. H., \& Kabtoul, M. (2026). eMEM: A hybrid spatio-temporal memory system for embodied agents. arXiv:2606.03374. \url{https://arxiv.org/abs/2606.03374}\par\vspace{1pt}

\par\noindent\hangindent=1.2em\hangafter=1 Shu, Y., Jonnalagedda, S. P., Gao, X., Jim\'{e}nez Guti\'{e}rrez, B., Qi, W., \& Das, K. (2026). REMem: Reasoning with episodic memory in language agents. arXiv:2602.13530. \url{https://arxiv.org/abs/2602.13530}\par\vspace{1pt}

\par\noindent\hangindent=1.2em\hangafter=1 Su, M., Guo, Y., Hou, Z., Bai, L., Li, Z., \& Zhang, Y. (2026). Beyond dialogue time: Temporal semantic memory for personalized LLM agents. In \textit{Findings of the Association for Computational Linguistics}. \url{https://doi.org/10.18653/v1/2026.findings-acl.1496}\par\vspace{1pt}

\par\noindent\hangindent=1.2em\hangafter=1 Tan, X., Wang, X., Liu, Q., Xu, X., Yuan, X., Zhu, L., \& Zhang, W. (2026). MemoTime: Memory-augmented temporal knowledge graph enhanced large language model reasoning. In \textit{Proceedings of the ACM Web Conference 2026} (pp. 4220--4231). \url{https://doi.org/10.1145/3774904.3792581}\par\vspace{1pt}

\par\noindent\hangindent=1.2em\hangafter=1 Vishwakarma, A., Lee, H., Suresh, M., Sharma, P. S., Vishwakarma, R., Gupta, S., \& Chauhan, Y. A. (2025). Cognitive weave: Synthesizing abstracted knowledge with a spatio-temporal resonance graph. arXiv:2506.08098. \url{https://arxiv.org/abs/2506.08098}\par\vspace{1pt}

\par\noindent\hangindent=1.2em\hangafter=1 Wan, F., Wu, C., \& Lyu, N. (2026). SodaMem: Evidence-grounded temporal graph memory for LLM agents. arXiv:2608.08055. \url{https://arxiv.org/abs/2608.08055}\par\vspace{1pt}

\par\noindent\hangindent=1.2em\hangafter=1 Wang, Z., Yu, B., Zhao, J., Sun, W., Hou, S., \& Liang, S. (2025). KARMA: Augmenting embodied AI agents with long-and-short term memory systems. In \textit{Proceedings of the IEEE International Conference on Robotics and Automation}. \url{https://doi.org/10.1109/ICRA55743.2025.11128047}\par\vspace{1pt}

\par\noindent\hangindent=1.2em\hangafter=1 Wu, D., Wang, H., Yu, W., Zhang, Y., Chang, K.-W., \& Yu, D. (2024). LongMemEval: Benchmarking chat assistants on long-term interactive memory. arXiv:2410.10813. \url{https://arxiv.org/abs/2410.10813}\par\vspace{1pt}

\par\noindent\hangindent=1.2em\hangafter=1 Yang, C., Zhou, C., Xiao, Y., Dong, S., Zhuang, L., \& Zhang, Y. (2026). Graph-based agent memory: Taxonomy, techniques, and applications. arXiv:2602.05665. \url{https://arxiv.org/abs/2602.05665} </content>\par\vspace{1pt}
\endgroup
\end{document}